\documentclass[letter,12pt]{article}
\usepackage{todonotes}
\usepackage{newtxtext,newtxmath, amsfonts, bm,mathtools}
\mathtoolsset{showonlyrefs=true}
\usepackage[utf8]{inputenc}
\usepackage[boxed,lined,linesnumbered]{algorithm2e}
\usepackage{amsmath, textcomp, paralist, bm, natbib}
\usepackage{array}
\usepackage{wrapfig}
\usepackage{multirow}
\usepackage{tabularx}
\usepackage{booktabs}
\usepackage{float, subfig}
\DeclareMathAlphabet{\mathcal}{OMS}{cmsy}{m}{n}
\graphicspath{{Graphics/}}
\usepackage{arydshln}
\usepackage{xcolor}
\usepackage[hyphens,spaces,obeyspaces]{url}
\usepackage{indentfirst}
\usepackage{pdflscape}
\usepackage[flushleft]{threeparttable}

\usepackage{pifont}

\usepackage[margin=3cm]{geometry}
\usepackage[margin=2cm]{caption}
\usepackage{titling}

\usepackage{tikz}
\usetikzlibrary{positioning}

\SetKwInOut{Parameter}{parameter}

\usepackage{multirow}

\usepackage{adjustbox}
\newcolumntype{R}[2]{%
    >{\adjustbox{angle=#1,lap=\width-(#2)}\bgroup}%
    l%
    <{\egroup}%
}

\usepackage{array, ragged2e}
\newcolumntype{P}[1]{>{\raggedright\arraybackslash}p{#1}}
\begin{document}

\title{Short-term prediction of stream turbidity using surrogate data and a meta-model approach}

\author{Bhargav Rele\textsuperscript{1}, Caleb Hogan\textsuperscript{2}, Sevvandi Kandanaarachchi\textsuperscript{2,3*$\dagger$} and Catherine Leigh\textsuperscript{1$\ddagger$} }
\date{ \scriptsize{ \today}}

\maketitle

\textsuperscript{1} Biosciences and Food Technology Discipline, School of Science, RMIT University, Bundoora VIC 3083, Australia. 

\textsuperscript{2} School of Science, Mathematical Sciences, RMIT University, Melbourne VIC 3000, Australia. 

\textsuperscript{3} CSIRO's Data61, Research Way, Clayton VIC 3168, Australia (Present address).

\textsuperscript{$\dagger$} ORCID: 0000-0002-0337-0395

\textsuperscript{$\ddagger$} ORCID: 0000-0003-4186-1678

* Corresponding author: sevvandi.kandanaarachchi@data61.csiro.au

\begin{abstract}

Many water-quality monitoring programs aim to measure turbidity to help guide effective management of waterways and catchments, yet distributing turbidity sensors throughout networks is typically cost prohibitive. To this end, we built and compared the ability of dynamic regression (ARIMA), long short-term memory neural nets (LSTM), and generalized additive models (GAM) to forecast stream turbidity one step ahead, using surrogate data from relatively low-cost in-situ sensors and publicly available databases. We iteratively trialled combinations of four surrogate covariates (rainfall, water level, air temperature and total global solar exposure) selecting a final model for each type that minimised the corrected Akaike Information Criterion. Cross-validation using a rolling time-window indicated that ARIMA, which included the rainfall and water-level covariates only, produced the most accurate predictions, followed closely by GAM, which included all four covariates. However, according to the no-free-lunch theorems in machine learning, no single model has an advantage over all other models for all instances. Therefore, we constructed a meta-model, trained on time-series features of turbidity, to take advantage of the strengths of each model over different time points and predict the best model (that with the lowest forecast error one-step prior) for each time step. The meta-model outperformed all other models, indicating that this methodology can yield high accuracy and may be a viable alternative to using measurements sourced directly from turbidity-sensors where costs prohibit their deployment and maintenance, and when predicting turbidity across the short term. Our findings also indicated that temperature and light-associated variables, for example underwater illuminance, may hold promise as cost-effective, high-frequency surrogates of turbidity, especially when combined with other covariates, like rainfall, that are typically measured at coarse levels of spatial resolution.

\end{abstract}



\section{Introduction}
Unnaturally high turbidity in rivers is a major aquatic ecosystem and human health concern, which makes it an important and commonly monitored water-quality variable in river management programs. High turbidity can indicate the presence of excess suspended sediment and particulate contaminants that can adversely affect water-dependent biota and ecosystem condition while complicating water treatment processes. Inorganic particles released by weathering affect the pH, alkalinity and metallicity of water while organic particles such as zooplankton and cyanobacteria can release toxins and odour components, potentially resulting in poor taste, smell and appearance, which together with contaminants from human, livestock and industrial waste can render water harmful for consumption \citep{Who2017} and increase water treatment  costs. Ecological processes such as bioturbation and human-induced factors, such as agricultural or industrial activities that accelerate erosion from catchments and increase sediment inputs to waterways, also contribute to turbidity problems \citep{leigh2019sediement}. Along with direct concerns for human health, excessive turbidity in freshwater ecosystems poses risks to aquatic organisms, for example by reducing light penetration and visibility and causing smothering, both in river systems themselves and in the receiving waters of downstream coastal and marine zones \citep{EventDrivenEcosystem}.

A prerequisite to informing management and policy aimed at preventing and responding effectively to the aforementioned phenomena, is the accurate and timely measurement and monitoring of turbidity. Turbidity is often measured in Nephelometric Turbidity Units (NTU) using $in$-$situ$ sensors that determine light scatter through water, which can be costly in terms of initial outlay, installation and maintenance \citep{lightattenuation, OnlineUV-VisSpectrophotometer}. Distributions of sensor networks require regular calibration and maintenance to reduce the likelihood of errors in measurement (technical anomalies) that may alert water management agencies incorrectly and potentially reduce public confidence in the data \citep{leigh2019framework}. Furthermore, high setup costs often mean sensors tend to be located sparsely across stream networks, typically at outlets in lower reaches, which decreases the ability to understand and manage water-quality dynamics across entire systems and spatially pinpoint turbidity issues in a timely fashion. 

A possible workaround is the development of models capable of predicting turbidity using high-quality data from covariates (sometimes referred to as surrogates). Such data can be acquired from lower-cost sensors installed at monitoring sites and/or from publicly available data sources, thereby reducing reliance on expensive sensors. Several types of models and surrogate variables have been explored and proposed in the literature, with rainfall and river discharge (hereafter flow) or water level commonly found as useful covariates (e.g. \cite{GMDHAlgorithms, leigh2019sediement}). Heavy rainfall affects turbidity via erosion and subsequent runoff, thereby increasing sediment loads in streamflow, which itself increases together with water level and acts to re-suspend sediments. Hence, sudden increase in turbidity often follows periods of high rainfall and sudden rise in water level \citep{leigh2019sediement}. However, turbidity may also increase when flow and water level decline leading to an increased concentration of particles \citep{iglesias2014turbidity}, although this phenomenon typically occurs more slowly than the increase in turbidity associated with sudden, fresh inputs of water. 

Seasonal patterns in temperature may also be reflective of seasonal patterns in rainfall, and thus flow and water level, such that interannual variation in air and water temperature may help to explain variation in turbidity. \cite{iglesias2014turbidity}, for example, found that water temperature was the most important variable for estimating river turbidity in northern Spain. Water temperature may also prove to be an informative covariate of turbidity given that suspended particles absorb more heat than water molecules when exposed to solar radiation \citep{Paaijmans2008}. As such we may expect that temperature correlates somewhat with turbidity, particularly during the day. This further suggests that sensors capable of recording underwater illuminance (the amount of light shining onto a surface, measured in lux) may help identify highly turbid waters given that poor water clarity will decrease light penetration. However, underwater illuminance will also be affected by the amount of solar irradiation and incident-light available, as dependent on, for example, time of day and year, cloud cover and shading from vegetation or other structures above the stream. As such, the relationship between light-associated variables (e.g. illuminance, irradiance, solar exposure) and turbidity, despite its potential utility, may not be as straight forward as that between rainfall, water level and turbidity. To our knowledge, the predictive ability of such light-associated variables as potential covariates of turbidity is yet to be explored beyond their potential use in determining light-attenuation \citep{TurbiditySensorriverMonitoring}, but is worth investigating given, for example, that low-cost $in-situ$ sensors that can measure both water temperature and illuminance are currently available (e.g. HOBO MX2202 data loggers; https://www.onsetcomp.com/).

Our first objective in this study was to assess the potential of variables that can be measured using low-cost $in$-$situ$ sensors or via readily available open-access databases (including rainfall, water level, temperature, and illuminance or radiant exposure, as available), either alone or in combination, to act as surrogates for the prediction of turbidity in rivers. We aim to achieve this by developing and comparing the ability of different models (dynamic regression, long short-term memory, and generalised additive models) to forecast the one-step ahead turbidity of Merri Creek, southeast Australia, using the aforementioned variables as covariates. Our second objective was to develop and implement the novel use of a ensemble machine learning method (meta-model) to select the most accurate model for turbidity prediction at any one time-step. The establishment of an accurate model has the benefit of providing water management and monitoring agencies a cost-effective tool for accurately predicting turbidity using surrogates. The use of surrogates may also serve to detect technical-anomalies within sensor-produced data via validation with model-produced data, enabling agencies to distinguish between turbidity fluctuations caused by natural phenomena and those caused by technical errors. By comparing an erratic datum in observed data to predicted data, erratic-behaviour not justified by the model can be flagged for investigation.  As such, potential benefits of using surrogate data provided by cheaper, readily available $in$-$situ$ sensors and/or publicly available data sources include increased reliability as well as cost-effectiveness. 

\section{Materials and Methods}

\subsection{Study area} 
Merri Creek is a roughly 70-km long tributary of the Yarra River, which flows through Wurrundjeri Country and the city of Melbourne, Australia, discharging into Port Phillip Bay. Water along some sections of Merri Creek is piped and some wetland areas have been drained and converted into channels or drains. Much of the catchment has been cleared of its native grassland and woodland, with land use changing in a downstream direction from rural to industrial to residential and urban. Monitoring turbidity is of interest to local government and community groups, particularly as the creek is home to vulnerable species such as the Growling Grass Frog ($Litoria raniformis$) and previously the platypus ($Ornithorhynchus anatinus$) for which elevated turbidity is considered a threatening process \citep{ConservationOfPlatypus, DELWP}.

\subsection{Data collection}
We sourced turbidity (NTU) and water level (m) data recorded hourly at Merri Creek between January 2013 and 2014 from the State of Victoria (Department of Environment, Land, Water and Planning) Water Measurement Information System  \citep{DptLandWaterPlan2014}. These data were recorded at a long-term monitoring site (St Georges Road, North Fitzroy) approximately 6 km upstream from the confluence with the Yarra River. Daily mean rainfall (mm), maximum air temperature ($^\circ$C) and total global solar exposure data (i.e. the total irradiance over a day, in  MJ/m$^{2}$) recorded at Australian Bureau of Meteorology weather stations that were closest to the North Fitzroy monitoring site \citep{Bom2014}. The rainfall data were recorded at Somerton Epping, while the temperature and solar exposure data were recorded at Melbourne Airport. Being only 20-25 km away, and in an upstream direction, from the monitoring site, rainfall, temperature and solar exposure data were expected to reflect phenomena occurring at the monitoring site. 

The temperature and solar exposure measurements are above-water measurements. Although measurements of temperature and light taken underwater would be useful, such data were unavailable due to unavoidable circumstances associated with the installation of new sensors at the monitoring site. Therefore, it was deemed reasonable to assume that, beyond the diel fluctuations in temperature and light being more buffered underwater, the Merri Creek water would be warmer on warmer days, given air temperature is a major factor controlling water temperature \citep{HydrologyResearch3no2}, and more light would reach and penetrate into the water column on sunnier days.  

\begin{figure}[!p]
    \centering
    \includegraphics[scale = 0.6]{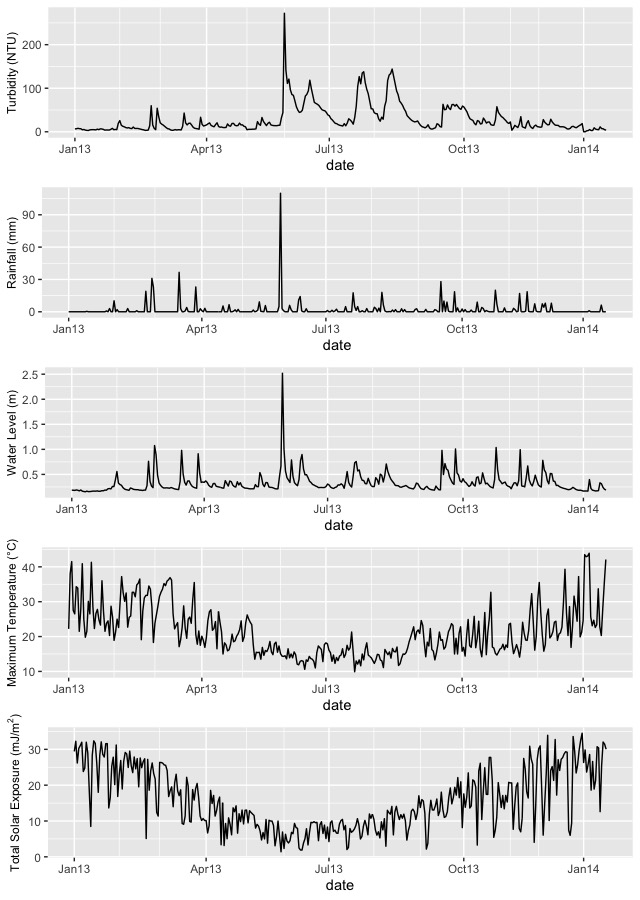}
    \caption{Mean daily turbidity, rainfall, water level, maximum air temperature and total global solar exposure during the study period.}
    \label{fig:exploratory}
\end{figure}

We then converted the hourly turbidity and water level data to daily means such that the final dataset contained daily turbidity, rainfall, water level, temperature and total global solar exposure data spanning January 2013 to January 2014 (n = 356 for each variable) (Figure \ref{fig:exploratory}).

\subsection{Modelling}

\subsubsection{Dynamic regression}
Time series regression models are often used to linearly predict dependent variables based on variation in single or multiple explanatory variables (covariates). The errors produced by such models are assumed to follow a white-noise process. In dynamic regression, dependent variables are predicted with the same principle in mind, however, with a key difference: the errors follow an auto-regressive integrated moving average (ARIMA) process. This allows errors from the regression to contain autocorrelations, an inherent feature of water-quality time series, and thus, account for hidden behaviour within the model \citep{HyndmanAthan}. 

To predict turbidity, we first built a dynamic regression model that included all four covariates (water level, rainfall, air temperature and solar exposure), then used a stepwise selection procedure to determine the best model (i.e. the best combination of covariates) as that which minimised the corrected Akaike Information Criterion (AICc). Goodness of fit was then evaluated using the Ljung Box test at the 5\% level of significance, and forecast results were assessed using residual diagnostics; ideally, the best model should not only have good fit (implying autocorrelations are not different from zero), but also have normally distributed residuals that in R Statistical Software \citep{RCoreTeam} using the packages \texttt{forecast} \citep{forecastR} and \texttt{fable}  \citep{OHaraWild2021}. 

\subsubsection{Long Short-Term Memory (LSTM)}
We also developed LSTM models (machine learning, neural network methods) given they are capable of learning long-term correlations in a sequence (e.g. associated with seasonality) and useful when analysing and predicting nonstationary processes, both features of water-quality time series \citep{rodriguez2020detecting}. Model parameters were specified in accordance with the LSTM model configurations used in \cite{LSTMNN}, given the similar context and performance capability requirements. Specifically, mean square error (MSE) was chosen as the loss function and adaptive moment estimation as the model optimization algorithm, the activation function was specified as the hyperbolic tangent to prevent the exploding gradient problem, the number of LSTM units was set to 10, and the number of epochs was set to 50 to avoid overfitting). Furthermore, a similar workflow to that used in \cite{LSTMNN} was utilised, which involved the following steps: data transformation, model building, model training; output production via point-by-point prediction. We used all four covariates to build the LSTM, using the \texttt{tslstm} \citep{PaulYeasin2022} package in R Statistical Software \citep{RCoreTeam}. 

\subsubsection{Generalized Additive Models (GAMs)}
Finally, we developed GAMs to predict turbidity given they are also capable of handling non-linear complex relationships and provide a good middle ground between linear regression models and their 'black-box', machine learning counterparts. Linear models, although relatively easy to implement and interpret, are incapable of explaining complex non-linear relationships between observations, while black-box models such as LSTMs that are able to handle such complexities are not straight-forward in finding the optimal result given they require tuning of multiple hyper-parameters. GAMs make it possible to fit flexible functions (capable of taking various shapes) onto the data. A GAM is fitted using a set of
basis functions, mostly splines, and the additive nature of GAMs combines these basis functions to create a non-linear smooth function that explains the relationship between variables. 
In this way, the smooth function is a combination of the non-linear basis functions, and the relationship between a predictor and the dependent variable is captured by multiple coefficients, in contrast to a single coefficient as in linear models. When this architecture is abstracted over all independent variables, the model will have produced a smoothing function that explains the relationship between each covariate and the response (in our case, turbidity). 

We used a stepwise selection procedure to determine the best smooth functions for each covariate's relationship with turbidity (rainfall, water level, solar exposure and air temperature) using splines of four, six and 12 degrees. The most appropriate smooth for each covariate relationship was selected as that which minimised the AICc (these were 4 degrees for each covariate except solar exposure, which was 12 degrees and suggestive of a more complex relationship between turbidity and solar exposure than between turbidity and each of the other covariates). We then introduced lag terms to the stepwise-selected model, specifically turbidity lagged at one and two observations prior to the response observation, to improve model accuracy. Lag terms were not specified as additional terms in the dynamic regression or LSTM models because these models automatically account for lags if needed. The validity of each covariate relationship (i.e. each smoothing function) was evaluated using ANOVA of parametric and non parametric effects at the 5\% level of statistical significance. If the non parametric effect is statistically significant, using a nonlinear smoothing function is justified (otherwise, a linear effect may be more suitable). Furthermore, forecast residuals were examined visually for their similarity to white-noise processes. GAMS were built using the \texttt{gam} (to determine the best smooth functions) and \texttt{mgcv} packages (to subsequently build and assess the model) in R statistical sorftware.

\subsection{Performance evaluation}
For each of the above model types (ARIMA, LSTM and GAM) cross-validation was carried out using a rolling-window approach, where each window consisted of 200 training observations and 1 testing observation. In the $1^{\text{st}}$ fold, covariate values from the first 200 observations were used to predict the turbidity of the $201^{\text{st}}$ observation. In the $2^{\text{nd}}$ fold, covariate values from the next 200 observations were used to predict the turbidity of the $202^{\text{nd}}$ observation, and so on. Hence, the cross-validation process consisted of 156 one-day-ahead predictions of turbidity (156 folds), each arising from an ARIMA model built on covariate data from the previous 200 days. Across all folds, the Root Mean Squared Error (RMSE) was then used to compare cross-validation results across each type of model. In general, a lower RMSE is preferred. 

\subsection{Meta-model}
\begin{figure}[!ht]
    \centering
    \includegraphics[scale=0.55,clip]{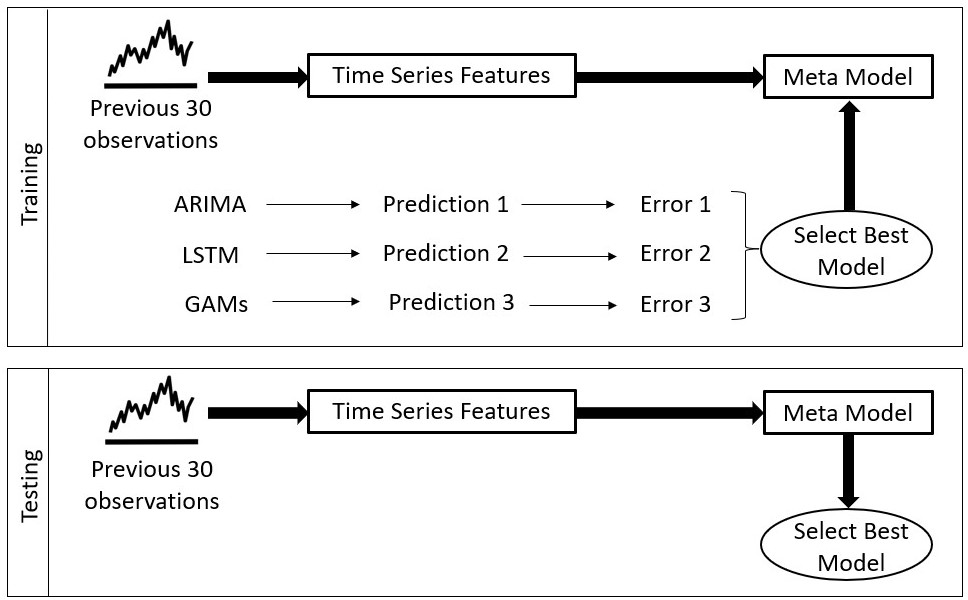}
    \caption{Schematic of the meta-model training (upper graphic) and testing process (lower graphic).}
    \label{fig:MetaModelDiagram}  
\end{figure}

In general, the term \textit{meta-model} means an integration of multiple sub-models, for example as implemented by ensemble machine learning methods, which are providing new advances in predictive capacity within the hydrological and water-quality modelling space \citep{MLHydrology}. In this study, we built an ensemble machine-learning meta-model to dynamically predict the best model, out of the ARIMA, LSTM and GAM options, at time $t$, without knowledge of the actual turbidity values at time $t$ (Figure \ref{fig:MetaModelDiagram}). 

In the training phase, we first extracted features from the turbidity time series, from $t-30$ to $t-1$, that described aspects of frequency, period, trend, spike, linearity, curvature, entropy, the autocorrelation function (acf) and extended acf, with the latter two features also considered from both the first and second differenced series, as a standard set of features considered when training such models. Using a machine learning method, such features can probabilistically determine the most suitable model for the next time step's prediction; the model that is probable to produce the lowest forecasting error can be selected to predict turbidity one-step ahead, thereby ensuring the most accurate model from all three models is used in each time period.

Predictions for the best ARIMA, LSTM and GAM models were used to train the meta-model. These predictions had been produced from the last 156 observations of the input time series (the first 200 observations having been used in the initial training fold such that predictions were not produced for those observations). Thus, to produce predictions from the meta-model and to compare results with each of the best individual models, we used a 30-day window prior to each prediction to compute the time-series features. In this way, we produced time-series features and predictions for 156 time points. Of these 156, we  used 70\% (110 observations) for training and the remaining 30\% (46 observations) for testing.

For each $t$, we used the predictions from ARIMA, LSTM and GAM to compute the absolute errors $e_t = |y_t - \hat{y}_t|$ for each model, where $y_t$ is the actual value and $\hat{y}_t$ is the predicted value. Let $e_{t,A}, e_{t,L}$ and $e_{t,G}$ denote the absolute errors at time $t$ for ARIMA, LSTM and GAM respectively. Thus, we defined the best model at time $t$ as that with the lowest absolute error:

 \begin{equation}\label{eq:meta1}
    \text{best\_model}(t) = 
     \begin{cases} 
      \text{ARIMA} & \text{if} \quad e_{t,A} = \min\{e_{t,A}, e_{t,L}, e_{t,G} \,   \} \\
      \text{LSTM} & \text{if} \quad e_{t,L} = \min\{e_{t,A}, e_{t,L}, e_{t,G} \,  \} \\
      \text{GAM} & \text{if} \quad e_{t,G} = \min\{e_{t,A}, e_{t,L}, e_{t,G} \,  \}
   \end{cases}
   \, .
\end{equation}

For each time window $[t-30, t-1]$, we then let $f(t)$ denote the extracted turbidity features, in order to produce $\left(f(t), \text{best\_model}(t) \right)$ for all time windows. We then used a random forest classifier as our meta-model, denoted by $\mathscr{M}$, to train on $f(t)$ and predict the $\text{best\_model}(t)$ using these data. We used random forests as the classifier because they are relatively easy to run, good fits are achievable without the need to tune parameters, overfitting is minimised \citep{RandomForest}, and they have shown promise for short-term water-quality prediction \citep{HybridDecisionTree}.

In the testing stage, we computed the turbidity time-series features as for training, and then used the trained meta-model to predict the best model from among the ARIMA, LSTM and GAM options. Suppose $f(t+1)$ denotes the time-series features for window $t+1$, then 
\begin{equation}
    \hat{b}\text{est\_model}(t+1) = \mathscr{M}(f(t+1))\, , 
\end{equation}
where $\hat{b}\text{est\_model}(t+1)$ is the predicted best model from the meta-model (Figure \ref{fig:MetaModelDiagram}). 

We used the R statistical software packages \texttt{tsfeatures} \citep{tsfeaturesR} and   \texttt{randomForest} \citep{randomForestRpackage} to produce and evaluate the meta-model. All individual models were run on the same machine (a MacBook Pro 12 with 1 processor, 2 cores, 2.7 GHz processor speed, and dual-core intel core i5) so that we could also compare processing times among them.

\section{Results}

\subsection{Dynamic regression (ARIMA)}
\begin{figure}[!ht]
    \centering
    \includegraphics[scale=0.8]{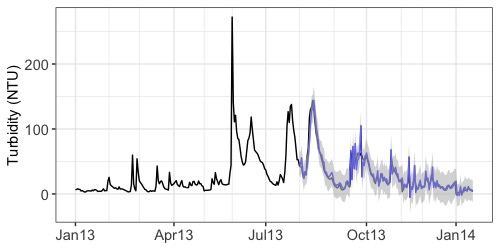}
    \caption{Turbidity (NTU) time series and cross validation results for one-step ahead forecasts over 156 folds, predicted using a dynamic regression model when rainfall, water level, solar exposure and air temperature are included as covariates and ARIMA(3,1,4) errors. Black line shows the observed turbidity times series. Blue line shows the turbidity forecasts, with grey-shaded 95\% confidence intervals.}
    \label{fig:cvARIMA}  
\end{figure}

The final dynamic regression model, having the lowest AICc, included rainfall and water level as covariates, with errors distributed as an ARIMA(3,1,4) process (Figure \ref{fig:cvARIMA}). Although solar exposure and air temperature were not included in this model, they were both included in the second and third best models, hinting that they (or their in-stream equivalents of light and water temperature) may still have predictive capabilities for turbidity in streams. The model had good fit in accordance with the Ljung-Box test carried out at the 5\% level of statistical significance, supporting the validity of the point forecasts and their confidence intervals that were constructed using the final model. Furthermore, the residuals followed an approximate white-noise process with no substantial autocorrelations up to lag 25.

In terms of performance, the model had reasonable accuracy when predicting turbidity one-step ahead, across all 156 folds of the rolling-window cross-validation procedure, as indicated for example by an RMSE of less than 9.5 NTU for the forecasts (Table~\ref{tab:forecastresults}) and the capture of both peaks and troughs in turbidity (Figure \ref{fig:cvARIMA}). The observed turbidity was well captured by the associated 95\% confidence interval (CI) of forecast turbidity, with the observed values exceeding the upper and lower bounds of the 95\% CI just 5\% of the time.

\begin{table}[!ht]
	\centering
	\caption{Forecast results of ARIMA, LSTM and GAM models. The root mean squared error (RMSE), mean absolute error (MAE), standard deviation (Std. Dev.) and error range for all three models. Errors are presented for one-step ahead forecasts made over 156 folds, in turbidity (NTU).}
	{
	\begin{tabular}{cp{2cm}p{2cm}p{2cm}p{3cm}}
		\toprule
   \ & RMSE & MAE & Std. Dev. & Error Range \\
        \midrule
ARIMA & \phantom{+}9.46  & \phantom{+}6.13  & \phantom{+}9.44  & (-42.37, 37.80) \\
LSTM  & 17.27 & 11.50 & 17.30 & (-44.01, 85.66) \\
GAM   & \phantom{+}9.89  & \phantom{+}5.91  & \phantom{+}9.86  & (-27.29, 52.51) \\
		 \bottomrule
	\end{tabular} }
	\label{tab:forecastresults}
\end{table}

\subsection{LSTM}
\begin{figure}[!ht]
    \centering
    \includegraphics[scale = 0.8]{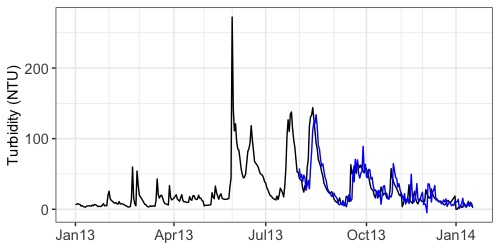}
    \caption{Turbidity (NTU) time series and cross validation results for one-step ahead forecasts over 156 folds, predicted using a LTSM model when rainfall, water level, solar exposure and air temperature are included as covariates. Black line shows the observed turbidity times series. Blue line shows the turbidity forecasts, with shaded 95\% confidence intervals.}
    \label{fig:cvLSTM}  
\end{figure}

The final LSTM model, like the final ARIMA model, appeared to capture both peaks and troughs in turbidity well (Figure \ref{fig:cvLSTM}). In contrast to the ARIMA model, however, the inclusion of solar exposure and temperature did not lead to an overestimation of turbidity values, such that the final LSTM model included all four covariates. Both the loss function (MSE) and the mean absolute error of the LSTM model stabilised after 10 epochs within each fold. 

The model had reasonable accuracy when predicting turbidity one-step ahead, across all 156 folds of the rolling-window cross validation procedure, although the forecasts' RMSE was almost double that of the ARIMA model at 17.3 NTU (Table~\ref{tab:forecastresults}). 

\subsection{Generalized additive model (GAM)}
The final GAM included the two lag terms for turbidity and four splines each for rainfall, water level and temperature. Although the step-wise selection procedure initially indicated 12 splines should be used for solar exposure, the AICc of the final model was minimised when the number of splines for this covariate were not specified. This was further supported by results of the ANOVA for parametric and non parametric effects; each smooth function except for solar exposure was significant at the 5\% level of statistical significance, which indicated the suitability of using nonlinear smooths to explain all covariate relationships bar that of solar exposure. However, excluding solar exposure from the model increased the model AICc and  RMSE of the forecasts. Hence, the final GAM included all four covariates along with the two lag terms for turbidity. Residuals for this final model closely resembled white noise, fluctuating predominantly around zero.  

\begin{figure}[!ht]
    \centering
    \includegraphics[scale=0.8]{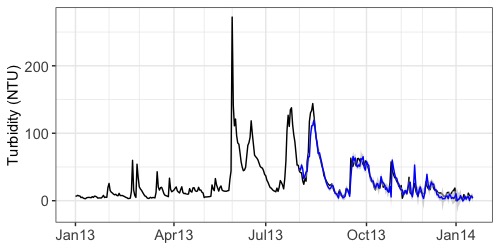}
    \caption{Turbidity (NTU) time series and cross validation results for one-step ahead forecasts over 156 folds, predicted using a GAM with rainfall, water level, solar exposure and air temperature as covariates. Black line shows the observed turbidity times series. Blue line shows the turbidity forecasts, with grey-shaded 95\% confidence intervals.}
    \label{fig:cvGAMs} 
\end{figure}

The final GAM performed similarly well to the ARIMA model, having a forecast RMSE of 9.9 NTU (when predicting turbidity one-step ahead across 156 folds of the rolling-window cross-validation procedure) (Table~\ref{tab:forecastresults}, Figure \ref{fig:cvGAMs}). However, given that the GAM had much narrower confidence intervals than those of the ARIMA (cf. Figure \ref{fig:cvARIMA}), there were times when the actual turbidity fell outside the GAM confidence intervals (Figure \ref{fig:cvGAMs})). 
For the majority of 1-period ahead forecasts, the associated 95\% confidence interval (CI) captured the observed turbidity value: 63\% of the observed values were captured within the upper and lower bounds of the 95\% CI, with 21\% exceeding the upper bound and 16\% exceeding the lower bound.

\subsection{Model comparison and the meta-model} 
\begin{figure}[!ht]
    \centering
    \includegraphics[scale=0.6]{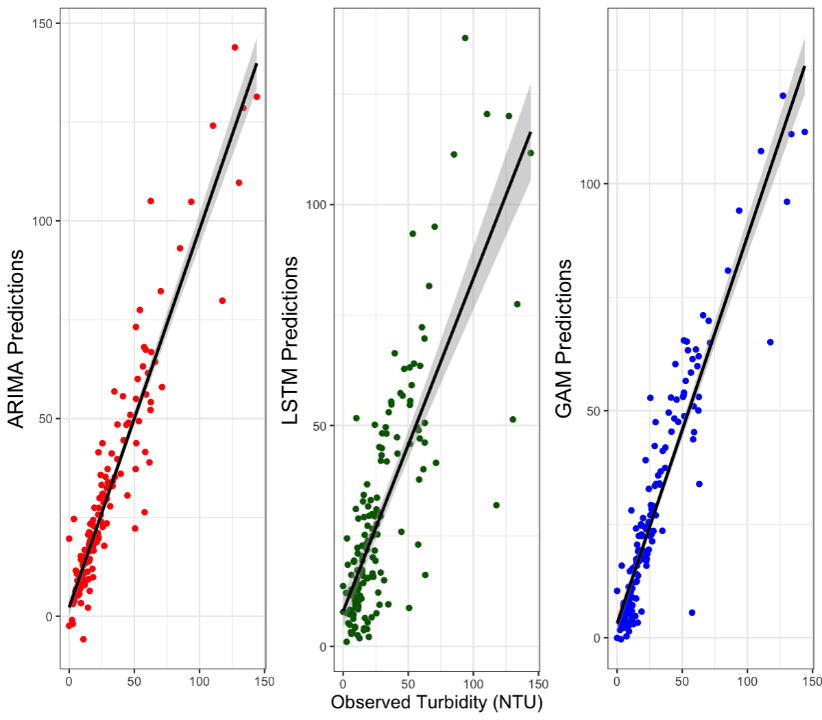}
    \caption{Scatterplot of response versus fitted values of the one-step ahead turbidity forecasts over 156 folds of the cross-validation procedure, for the final ARIMA, LSTM and GAM. Grey shading shows the standard errors. }
    \label{fig:ObVSPredCompressed}  
\end{figure}

\begin{figure}[!ht]
    \centering
    \includegraphics[scale=0.9]{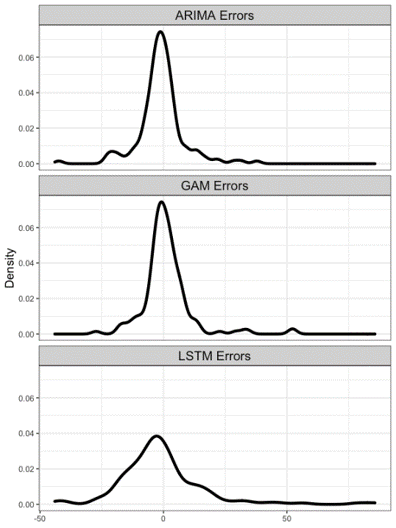}
    \caption{Density plot of forecast errors produced by the final ARIMA, LSTM and GAM (one-step ahead forecasts over 156 folds of the cross-validation procedure)}
    \label{fig:DensityErrorsCompressed}  
\end{figure}

Our findings suggested that the final ARIMA outperformed the GAM and LSTM models. For example, the final ARIMA had the lowest forecast RMSE and standard deviation among all three model types, followed closely by the GAM, which had the lowest forecast MAE. Visual inspection of plots of observed versed predicted turbidity values also showed that the ARIMA (and the GAM) had lower variance overall than the LSTM (Figure \ref{fig:ObVSPredCompressed}), and the density distributions of forecast errors for all three models approached normality (with ARIMA and the GAM errors having higher kurtosis than LSTM errors; Figure \ref{fig:DensityErrorsCompressed}). Furthermore, a greater proportion of the observed turbidity values was captured by the 95\% CI of the ARIMA forecasts than by  the 95\% CI of the GAM forecast. Finally, the LSTM required more information and processing time than either of the two other models, taking approximately 150 to 350 times longer to run than the ARIMA and GAM, respectively.

However, the meta-model provided a more nuanced ability to accurately predict turbidity by selecting the optimal model (out of ARIMA, LSTM and GAM) at each time step, selecting the ARIMA, LSTM and GAM 39.1\%, 21.7\% and 39.1\% of the time, respectively. 
The meta-model outperformed all other models, including the ARIMA, having the lowest RMSE, MAE, standard deviation and the narrowest forecast error range of all models (Table~\ref{tab:metamodelresults}).  

\begin{table}[!ht]
	\centering
	\caption{Comparison of meta-model forecast results with those of ARIMA, LSTM and GAM. The root mean squared error (RMSE), mean absolute error (MAE), standard deviation (Std. Dev.) and error range for all three models. Errors are presented for one-step ahead forecasts made over 156 folds, in turbidity
(NTU)}
	{
	\begin{tabular}{cp{2cm}p{2cm}p{2cm}p{2.5cm}}
		\toprule
   \ & RMSE & MAE & Std. Dev. & Error Range  \\
        \midrule
 ARIMA              & 5.44  & 3.66 & 8.62 & (0.001, 19.62) \\
 LSTM               & 8.72  & 6.53 & 8.54 & (0.25, 21.46)  \\
 GAM               & 5.93  & 4.50 & 8.70 & (0.01, 17.17)  \\
Meta-model          & 2.31  & 1.67 & 6.60 & (0.001, 7.80)  \\
		 \bottomrule
	\end{tabular} 
	}
	\label{tab:metamodelresults}
\end{table}

\section{Discussion}

Our two main objectives for this study were to (i) explore the potential of open-access and/or easily measured and cost-effective environmental and hydrological variables (rainfall, water level, solar exposure and air temperature) to act as surrogates of turbidity in rivers, and (ii) use a novel meta-model approach to select the most accurate model for the short-term forecasting of turbidity at any one time step. In terms of our first objective, we found that rainfall and water level were consistently useful as covariates when predicting turbidity one-step ahead, being included in all three final models (i.e. the dynamic regression (ARIMA), long short-term memory (LSTM) and generalised additive models (GAM)). Changes in turbidity often reflect changes in rainfall and water level \citep{leigh2019sediement} due to the hydrogeomorphological processes involved in sediment transport to and along streams, so our finding that these two variables can act as surrogates of turbidity therefore supports such a theoretical explanation. Furthermore, the model providing the most accurate predictions, on average, of one-step ahead turbidity (the ARIMA, followed closely by the GAM) included the rainfall and water-level covariates only. 

We were less certain about the potential of solar exposure and air temperature to act as turbidity surrogates, given the relationships between these two variables and turbidity are less direct and more complex than relationships between turbidity, rainfall and water level. Despite this, both solar exposure and air temperature were included in both the final GAM and LSTM models, each of which provided reasonable turbidity forecasts, with the GAM performing almost as well as the ARIMA. Given that underwater light illuminance and water temperature are heavily influenced by solar exposure and air temperature and are easily measured at high-frequency by relatively low-cost in-situ sensors, our results suggest that the former two covariates also hold promise as surrogates for turbidity,  particularly when combined with other covariates such as rainfall that are typically measured at coarser spatial and/or temporal scales (e.g. daily in the present study). 

Not only was turbidity most accurately forecast by the ARIMA but the observed values of turbidity were also more often captured within the 95\% CIs of the ARIMA than they were within those of the GAM. We were unable to compute confidence intervals for the LSTM; confidence and prediction intervals in deep learning are both developing, active areas of research \citep{Du2022, Sun2021} and the software packages currently implementing LSTM (e.g. \texttt{tslstm}) do not yet provide the functionality to calculate these intervals, partly because there is no consensus on which method is most suitable. While we acknowledge this limitation of our model comparison, all the performance evaluation metrics that we were able to calculate pointed to ARIMA being the most accurate model of the three trialled, at least on average. Indeed there were some specific time steps when one or both of the latter two models had higher accuracy than the ARIMA model. Our meta-model results supported this proposition, given that it outperformed the ARIMA (indeed all three final models) in terms of its ability to accurately predict turbidity one-step ahead by alternating among the three models for different time points. To this end, we met our second objective of the study, with our findings demonstrating the benefit of using such an ensemble approach for turbidity forecasting and indicating that such an approach may be a viable alternative to using measurements sourced directly from turbidity-sensors where costs prohibit their deployment and maintenance, and when predicting turbidity across the short term.

While the use of meta-model approaches in water-quality applications is not completely new or consistent, our work progresses this active area of research by providing the first example, to our knowledge, of a meta-model that selects the best sub-model at each time step for the short-term prediction of turbidity in rivers. In contrast to our approach, water-quality studies that describe \lq{meta-models\rq} appear to either combine multiple model components to formulate an overarching model (rather than selecting the best sub-model at each instance) or use a machine learning model as a surrogate for a process-based model. For example, \cite{Masoumi2022} proposed a water quality-quantity model using a surrogate neural network that emulated a process-based, two-dimensional hydro-dynamics model. \cite{Holzkamper2012} used a Bayesian network approach to produce an integrated \lq{meta-model\rq} from loose numerical and knowledge based sub-models that each specialised in a different aspect of catchment management, such as flood risk, land and sewage management. \cite{Lerios2019} used multiple machine learning algorithms to predict a state-based index of water quality and to choose an overall \lq{best\rq} model rather than constructing a meta-model using all algorithms. Similarly, \cite{HybridDecisionTree} used multiple machine learning models for short-term water-quality prediction but did not use a meta-model to combine the strengths of each individual model. 

Our use of the term \lq{meta-model\rq} is similar in sentiment to the research conducted by \cite{Nasir2022}, in which several machine learning classifiers were used to predict a state-based water-quality index, and to train a meta-model that selected the best sub-model for each instance. While they too found that their meta-model increased the prediction accuracy, the problem being addressed by our meta-model differs substantially from theirs: \cite{Nasir2022} addressed a non-time dependent classification problem (predicting a categorical water-quality index, in a similar fashion to \cite{BUI2020137612}) whereas we address a time-dependent regression problem (predicting turbidity values through time). 

Finally, our study advances upon the work of \cite{LSTMNN} and \cite{leigh2019framework}, in which LSTM and ARIMA models were used respectively to predict turbidity data using covariates for the purposes of technical-anomaly detection, and thus further showcases the capabilities of such models in forecasting water-quality variables. For example, the ARIMA and LSTM (and GAM) models, via implementation in a meta-model, could be used to more accurately flag sensor-produced turbidity measurements that deviate substantially from predicted turbidity values and potentially then be corrected as part of a quality control and assurance protocol. We also suggest that our meta-model approach could be further improved by considering (i) a larger collection of sub-models (e.g. XGBoost, Exponential Smoothing;  \cite{DEB2017902}) and (ii) a larger collection of potential meta-model candidates. We selected a random forest classifier because it does not over-fit the data and is considered a robust, ensemble classifier as it uses multiple decision trees under the hood; however, future studies may wish to trial other classifiers such as SVMs, XGBoost or neural networks. Most importantly, our study has demonstrated that a meta-model approach can be used to accurately predict turbidity based on cost-effective surrogate data, which has the potential to benefit water management and monitoring agencies, along with human and freshwater ecosystem health via the timely pinpointing and prediction of turbidity events in river networks.

\section*{Supporting Information}
All programming scripts and high resolution figures are available at \url{https://github.com/sevvandi/supplementary_material/tree/master/MerriCreek}.

\section*{Acknowledgements}
Merri Creek flows through country of the Wurundjeri people of the Eastern Kulin Nation. We acknowledge the Traditional Owners and ongoing Custodians of these unceded lands and waters, and the significance of Merri Creek to the Wurundjeri people. Funding support for this study was provided an RMIT University Information Systems (Engineering) Enabling Capabilities Platform `2020 Capability Development Fund' grant, and an Australian Research Council Centre of Excellence for Mathematical and Statistical Frontiers (ACEMS; ARC grant number CE140100049) `Research Sprint Scheme' grant. We thank  City of Whittlesea, City of Moreland, and the Merri Creek Management Committee for engaging discussions and their support of this study. Solar exposure, air temperature and rainfall data were provided by the Australian Bureau of Meteorology and the turbidity and water-level data were provided by the State of Victoria (Department of Environment, Land, Water and Planning) Water Measurement Information System.

\bibliography{references}

\bibliographystyle{agsm}

\end{document}